\pdfoutput=1

\documentclass[11pt]{article}

\usepackage[preprint]{acl}

\usepackage{times}
\usepackage{latexsym}

\usepackage[T1]{fontenc}

\usepackage[utf8]{inputenc}

\usepackage{microtype}

\usepackage{inconsolata}

\usepackage{graphicx}

\usepackage{url}
\usepackage{booktabs}
\usepackage{multirow}
\usepackage{array}
\usepackage{makecell}
\usepackage{diagbox}

\usepackage{amsmath}
\usepackage{algorithm}
\usepackage{algorithmic}
\usepackage{newfloat}
\usepackage{listings}
\DeclareCaptionStyle{ruled}{labelfont=normalfont,labelsep=colon,strut=off} 
\floatstyle{ruled}
\newfloat{listing}{tb}{lst}{}
\floatname{listing}{Listing}

\title{DCR: Quantifying Data Contamination in LLMs Evaluation}

\author{
Cheng Xu\textsuperscript{1,3} \quad 
Nan Yan\textsuperscript{2,3} \quad 
Shuhao Guan\textsuperscript{1} \quad 
Changhong Jin\textsuperscript{1} \\ 
\textbf{Yuke Mei\textsuperscript{3} \quad 
Yibing Guo\textsuperscript{1} \quad 
M-Tahar Kechadi\textsuperscript{1}} \\
  \textsuperscript{1}\,University College Dublin \quad 
  \textsuperscript{2}\,Georgia Institute of Technology \quad 
  \textsuperscript{3}\,Bebxy\\
  \texttt{cheng.xu1@ucdconnect.ie \quad tahar.kechadi@ucd.ie} \\
}

\begin{document}
\maketitle
\begin{abstract}
The rapid advancement of large language models (LLMs) has heightened concerns about benchmark data contamination (BDC), where models inadvertently memorize evaluation data during the training process, inflating performance metrics, and undermining genuine generalization assessment. This paper introduces the Data Contamination Risk (DCR\footnote{\url{https://github.com/chengxuphd/dcr}}) framework, a lightweight, interpretable pipeline designed to detect and quantify BDC risk across four granular levels: semantic, informational, data, and label. By synthesizing contamination scores via a fuzzy inference system, DCR produces a unified DCR Factor that adjusts raw accuracy to reflect contamination-aware performance. Validated on 9 LLMs (0.5B-72B) across sentiment analysis, fake news detection, and arithmetic reasoning tasks, the DCR framework reliably diagnoses contamination severity and with accuracy adjusted using the DCR Factor to within 4\% average error across the three benchmarks compared to the uncontaminated baseline. Emphasizing computational efficiency and transparency, DCR provides a practical tool for integrating contamination assessment into routine evaluations, fostering fairer comparisons and enhancing the credibility of LLM benchmarking practices.
\end{abstract}

\section{Introduction}

Large Language Models (LLMs) have demonstrated exceptional performance across a wide range of Natural Language Processing (NLP) tasks. This success is largely attributed to their pre-training on vast amounts of data, followed by task-specific fine-tuning \citep{zhao2023survey}. Typically, LLM performance is evaluated using standard benchmarks designed to measure performance across various tasks \citep{chang2024survey}. However, as LLMs grow in size and are trained on increasingly extensive corpora, an important challenge emerges: Benchmark Data Contamination (BDC) \citep{xu2024benchmark,xu2025ssa}. This occurs when content related to benchmark datasets is included in the model's pre-training phase, either through direct inclusion of the benchmark data or by exposure to contextual information surrounding these benchmarks \citep{sun2025the}. Such contamination can lead to skewed evaluations and limiting the ability to accurately assess real generalization capabilities \citep{xu2025triplefact}. The increasing body of research recognizes that addressing this issue is critical for ensuring the reliability and fairness of LLM evaluations \citep{lee2022deduplicating,sainz2023nlp,zhou2023dont,mcintosh2024inadequacies,deng2024unveiling}.

Detecting and mitigating BDC is a complex challenge, particularly in the scenario that many of the state-of-the-art LLMs are closed-source commercial products, with proprietary architectures and training methodologies that remain inaccessible to the public \citep{openai2024gpt4,anthropic2024introducing,geminiteam2024gemini}. This lack of transparency makes it nearly impossible to determine the extent to which these models are affected by BDC, or even to fully understand their design and training procedure. However, in the case of open-source LLMs, the situation is not much better. Although model architectures and weights are publicly available, the pre-training corpus are often undisclosed \citep{touvron2023llama,jiang2023mistral}. This opacity raises concerns about the credibility of performance claims made in papers and technical reports.

Recent efforts to detect BDC have employed a variety of techniques, ranging from direct matching-based methods, such as n-gram matching \citep{balloccu2024leak,openai2024gpt4,ippolito2023preventing,brown2020language} and pre-training corpus membership inference \citep{li2024task,shi2024detecting,chang2023speak,duarte2024decop,ye2024data}, to more sophisticated comparison-based approaches by analyzing the behavior of LLMs on benchmark tasks, using metrics such as similarity \citep{riddell2024quantifying}, preference \citep{golchin2024data}, perplexity \citep{li2023estimating}, chronological order \citep{golchin2024time,roberts2023data,yang2023rethinking}, sequence alignment \citep{kandpal2022deduplicating,oren2024proving}, and prediction distribution \citep{magar2022data,shi2024detecting}. Although these approaches have shown promise, they face several significant challenges that many are computationally expensive, requiring substantial resources to execute at scale, and often lack interpretability, making it difficult to draw clear conclusions or apply them effectively during the development of LLMs \citep{xu2024benchmark}. These limitations hinder the broader adoption of BDC detection methods in practical settings, highlighting the need for more efficient and transparent techniques to ensure the integrity of LLM evaluations.

In this paper, we present a novel and light framework for detecting BDC in LLMs, called the \textbf{D}ata \textbf{C}ontamination \textbf{R}isk (DCR) framework. Designed to address key limitations in existing methods, the DCR framework emphasizes both interpretability and computational efficiency, making it suitable for real-world applications with limited resources. Our main \textbf{contributions} are as follows: (1) we introduce the DCR framework, a more interpretable and resource-efficient approach to BDC detection; (2) we propose a new metric, the DCR Factor, to quantify the contamination risk of LLMs and benchmarks. This metric not only identifies potential BDC but also provides a mechanism to adjust performance results on contaminated benchmarks; (3) We extensively conducted tests of three benchmarks on 9 LLMs (0.5-72B parameters) using the DCR framework to understand their BDC risks.

\section{DCR Framework} 
\label{sec:framework}

\subsection{Motivation}
\label{sec:motivation}

With the rapid rise of LLMs, research on BDC has gained significant attention, leading to increasingly sophisticated and fancy approaches. However, an intriguing phenomenon has emerged: despite the growing body of work on BDC detection and mitigation, few of these methods have been integrated into the actual LLM development process. Instead, much of the research seems to operate in isolation, without practical impact - a situation that undermines its value, which is not what we want to see.

Through a detailed investigation into LLM development practices and current BDC research, we have identified a primary reason for this disconnect: the enormous scale of pre-training corpora, often exceeding one trillion tokens \citep{touvron2023llama,openai2024gpt4,glm2024chatglm,deepseekai2025deepseekr1}, and even involves steps like distillation \citep{xu2024survey} and RLHF \citep{kaufmann2024survey}. Processing such large datasets using complex BDC detection or mitigation techniques demands extensive computational resources, and crucially, does not result in substantial improvements in the final model performance. As a result, developers are reluctant to invest heavily in BDC prevention, typically opting for simpler and less resource-intensive methods, such as n-gram matching \citep{touvron2023llama2}.

Additionally, even when state-of-the-art BDC detection or mitigation techniques are applied, they cannot fully guarantee a contamination-free model, especially in cases of more nuanced, semantic level BDC \citep{xu2025ssa}. Furthermore, the effectiveness of BDC mitigation is difficult to quantify, making it hard to justify the considerable resource expenditure. In essence, allocating substantial computational resources to ensure perfect evaluation integrity is often seen as an effort with diminishing returns, offering little practical benefit to model performance.

When designing the DCR evaluation framework, we primary focus on optimizing efficiency, aim to assess the BDC risk using minimal test cases while still maintaining the reliability and validity of the evaluation results. To facilitate this, we introduced a novel metric, the DCR Factor, which quantifies BDC risk more intuitively during benchmark evaluations. Another key consideration was interpretability. Rather than relying on abstract metrics such as perplexity \citep{li2023estimating} or model preferences \citep{li2025preference}, which often operate as black-box indicators \citep{fang2025what,hu2024can}, we sought to provide clear, easily understandable explanations for BDC risk. The DCR framework offers transparent insights into the BDC risks, allowing developers to grasp the implications of the results directly.

Our ultimate goal is to encourage LLM developers to integrate BDC evaluation into standard benchmark testing. By doing so, a more objective and reliable result can be provided to ensure that the performance metrics of LLMs are accurate and contamination-aware. We believe that this approach will improve the integrity of LLM evaluations in the field.

\subsection{Problem Definitions}

In this section, we formulate the problem of BDC and its detection, characterize its different levels, and outline the challenges it poses.

Let $D_{train}$ be the pre-training dataset of LLMs, consisting of a large corpus of text; $B$ be the benchmark dataset used for evaluating the performance of LLMs; $D_{train}^{info}$ and $B^{info}$ be the sets of all information in $D_{train}$ and $B$, respectively.

\textbf{Benchmark Data Contamination} occurs when there is an overlap between $D_{train}$ and $B$, either directly or indirectly, leading the model to have prior exposure to the evaluation data or related knowledge. This overlap may not be detectable through direct token matches, but can significantly impact the model evaluation process. Formally, BDC risk can be quantified by the following equation:

\begin{equation}
\label{eq:1}
    \text {BDC}=\frac{|D_{train}^{info} \cap B^{info} |}{|B^{info} |}
\end{equation}

However, in order to accurately assess the BDC risk during the evaluation of LLMs, it is clear that Equation \ref{eq:1} is too general, and a more fine-grained system for quantifying BDC needs to be established. To this end, we adopt the work of \citet{xu2024benchmark}, who categorize BDC into four levels: \textit{Semantic Level} (L1), \textit{Information Level} (L2), \textit{Data Level} (L3) and \textit{Label Level} (L4). For example, imagine an LLM as a student about to take a final exam. Semantic level BDC occurs when students receive exam questions in different wording before the test. Information level BDC happens when the student obtains information beforehand, such as "over 60\% of the answers to the multiple-choice questions are option C" or "the exam content focuses on Chapter 3 of the textbook." Data level BDC involves students getting the actual exam questions in advance, while label level BDC involves having both the questions and the correct answers beforehand. All these factors can influence the student's performance in varying degrees. 

Based on this system, BDC can be quantified and defined more systematically and at a more fine-grained level, detailed definitions and examples of the four BDC levels are provided in Appendix \ref{sec:4level}.

\subsection{Objectives}
Given a pre-trained LLM $M$, its pre-training corpus $D_{train}$, and a benchmark dataset $B$, our goal is to:

\begin{enumerate}
    \item \textbf{Detect BDC:} Identify the presence of contamination at each BDC level $L_i, i\in \{1, 2, 3, 4\}$.
    \item \textbf{Quantify BDC:} Compute contamination scores $S_i$ for each level $L_i$, reflecting overlap on various dimensions between $D_{train}$ and $B$.
    \item \textbf{Adjust Evaluation Metrics:} Adjust the performance metrics of LLMs to account for the detected BDC risk, ensuring a more accurate assessment of its generalization capabilities.
\end{enumerate}

\begin{figure*}[t]
    \centering
    \includegraphics[width=1\linewidth]{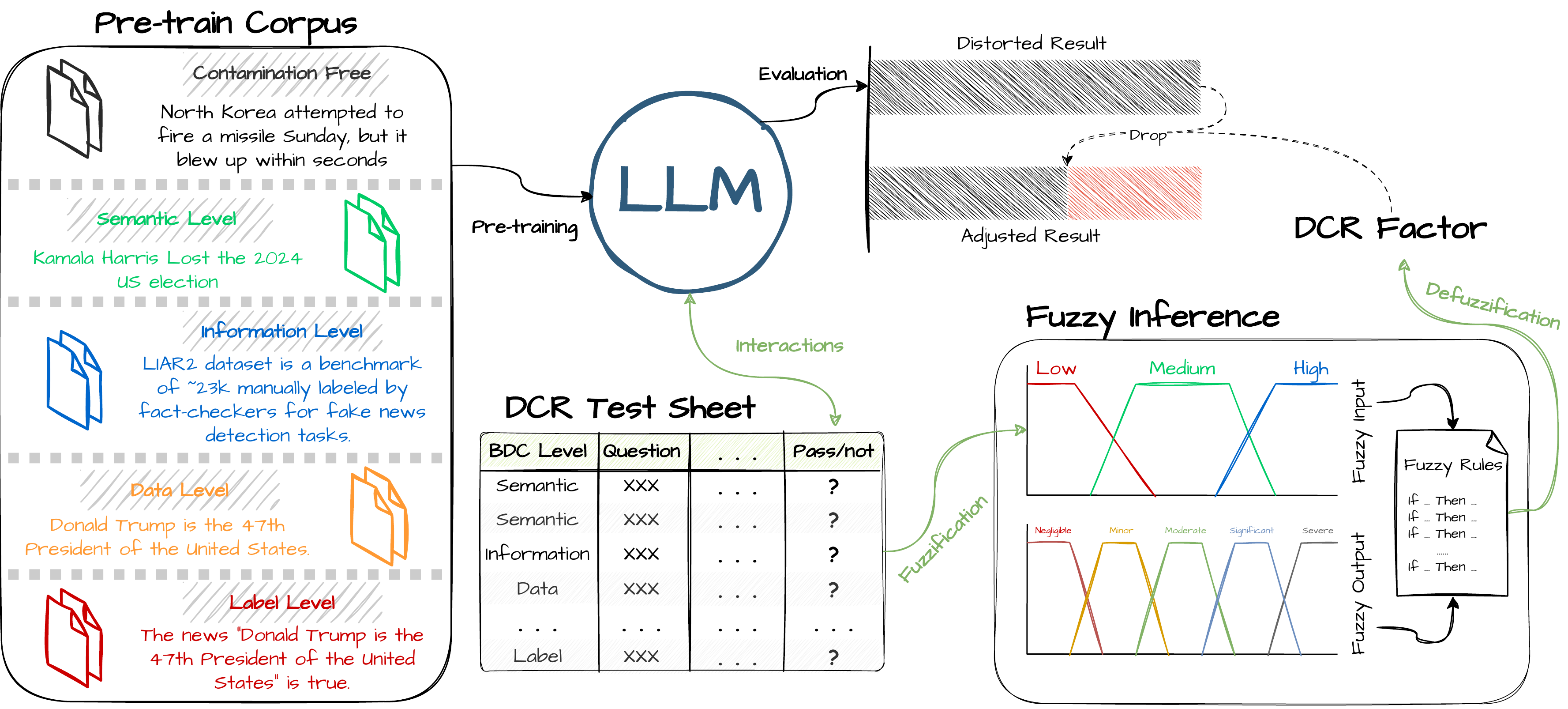}
    \caption{The Data Contamination Risk (DCR) evaluation framework diagram.}
    \label{fig:dcr}
\end{figure*}

\subsection{DCR Workflow}

The DCR framework provides an efficient method to detect and quantify the BDC risk during the LLMs evaluations. The DCR framework is structured into two key stages: the \textit{Quantification Stage} and the \textit{Adjustment Stage}, depicted in Figure \ref{fig:dcr}.

\subsubsection{Quantification Stage}
\label{sec:quantification}

The quantification stage focuses on detecting and quantifying contamination risk across predefined BDC levels. Central to this process is the \textbf{DCR Test Sheet}, a structured set of prompts designed to evaluate contamination risks at different levels, denoted as $L_i$.

For each contamination level $L_i$, a contamination score $S_i$ is calculated based on the responses $R_i$ of the model to a series of test prompts $P_i$. The contamination score is defined as:
\begin{equation}
\label{eq:2}
    S_{i} =\frac{1}{N_{i}}\sum _{k=1}^{N_{i}} Check\left( p_{i}^{k} ,r_{i}^{k}\right)
\end{equation}
where $p_{i}^{k} \in P_{i}$ and $r_{i}^{k} \in R_{i}$ represent specific prompt-response pairs. $N_i$ is the number of test prompts for contamination level $L_i$, and the function $check$ determines whether the response $r_{i}^{k}$ exhibits contamination based on the prompt $p_{i}^{k}$.

Ideally, test prompts $P_i$ should be provided by the benchmark creators to ensure relevance and consistency. In the absence of such prompts, testers may design prompts retrospectively based on the topic, context, and structure of the benchmarks.

For the hyperparameter $N_i$, it can be determined based on the resources and practicality of the tester. Intuitively, one would assume that increasing the size of $N_i$, i.e., increasing the amount of testing, would result in less error. A more in-depth investigation of $N_i$ is presented in Appendix \ref{sec:numberOfTest}.

For the \textit{Check} step, our default approach is to do this manually and in a binary presentation of the results, since the purpose of the DCR framework is to complete the assessment of the BDC risk with as few resources as possible, so that the tester can quickly record the results based on the responses of the model when collecting or organizing the test prompts. Certainly, the $check$ step can be replaced by another model, e.g., BERT \citep{devlin2019bert}, and the presentation of results can also be changed from discrete to continuous. We provide a nuanced assessment of the use of models as an alternative to human labor for the \textit{Check} step in Appendix \ref{sec:check}.

\subsubsection{Adjustment Stage}
\label{sec:adjustment}

The Adjustment Stage of the DCR framework focuses on calibrating evaluation results of LLMs by accounting for identified BDC risk. After quantifying contamination scores $S_{i}$, this stage employs a fuzzy inference system to compute a comprehensive \textbf{DCR Factor}, subsequently adjusting the raw performance metrics to reflect a more accurate assessment of model generalization capabilities.

To achieve this, we established a fuzzy inference system incorporating defined membership functions and inference rules designed to interpret contamination scores in a nuanced yet interpretable manner. The reason why we chose to use fuzzy logic for the DCR Factor calculation is because the definition of the BDC level itself is fuzzy, so it is very appropriate and explainable to simulate the process \citep{das2020survey,zadeh1965fuzzy,zadeh2023fuzzy}. The fuzzy logic system consists of input variables representing contamination scores at the semantic, information, data, and label levels (denoted as $S_1, S_2, S_3, S_4$), each ranging between 0 and 1. We then defined membership functions for three contamination degrees: \textit{Low}, \textit{Medium}, and \textit{High}. The output variable, the DCR Factor, similarly ranges from 0 to 1, quantified by membership functions representing contamination severity: \textit{Negligible}, \textit{Minor}, \textit{Moderate}, \textit{Significant}, and \textit{Severe}.

Formally, each input score $S_i$ is fuzzified into membership degrees $\mu_{Low}(S_i)$, $\mu_{Medium}(S_i)$, and $\mu_{High}(S_i)$:
\begin{equation}
\mu_{term}(S_i) = membership(S_i, [a, b,...])
\end{equation}
where $membership$ denotes the membership function defined by parameters $[a, b,...]$, for example trapezoidal membership function.

The adjustment process relies on several fuzzy inference rules that intuitively represent real-world contamination impacts, for example: \textit{High} contamination at the Label Level (L4) indicates a \textit{Severe} risk, significantly impacting the integrity of evaluation results. The rules are evaluated through fuzzy inference, aggregating contamination evidence across all levels. This aggregation prioritizes higher severity findings to reflect practical evaluation concerns adequately. Detailed settings of fuzzy inference system are provided in Appendix \ref{sec:fuzzy}.

The final DCR Factor is obtained by defuzzifying the aggregated fuzzy set, yielding a crisp numerical value indicative of the contamination risk. Subsequently, this factor is applied to adjust the original accuracy metric (Acc), ensuring that reported model performance realistically accounts for the contamination risk:
\begin{equation}
    \text {Acc}_{adj} = \text {Acc} \times (1 - \text {DCR\ Factor})
\end{equation}

This adjusted accuracy (Acc$_{adj}$) provides a contamination-aware performance metric, significantly enhancing the reliability and fairness of benchmark evaluations. \textit{It is also important to emphasize that we do not consider this adjusted accuracy to be the accurate performance of the model after removing all BDC risk, but rather a result that incorporates BDC risk considerations.}

\section{Experiments}
\label{sec:experiments}

The main thrust of this section is to demonstrate that the DCR framework is capable of detecting and quantifying BDC risk at a low computational cost. Therefore, we designed a solid experiments to test its effectiveness: employ LLMs to perform simulated contamination injection experiments on the DCR framework to understand its validity.

\subsection{Experiment Setup}
\label{sec:injection_exp}

\begin{table}[b]
    \centering
    \resizebox{0.48\textwidth}{!}{%
    \begin{tabular}{l|c|c|c}
    \toprule
        \textbf{Benchmark} & SST-2 & LIAR2 & GSM8K \\
        \midrule
        \textbf{Task Type} & Sentiment Analysis & Fake News Detection & Arithmetic Reasoning \\
        \midrule
        \textbf{Release Year} & 2013 & 2024 & 2021 \\
        \midrule
        \textbf{Subset} & train/val/test & train/val/test & train/test \\
        \midrule
        \textbf{Size} & 67.3k/872/1.82k & 18.4k/2.3k/2.3k & 7.47k/1.32k \\
        \midrule
        \textbf{Label Type} & Binary & 6-Class & Open-ended \\
    \bottomrule
    \end{tabular}
    }
    \caption{Overview of the three benchmarks used in the experiments.}
    \label{tab:benchmarks}
\end{table}

To study more clearly the impact of BDC on the benchmark results of language models, we follow the work of \citet{jiang2024investigating} and \citet{bordt2025forget}, use the similar procedure to perform contamination injection experiments. In this experiment, we will inject benchmark-relevant contaminated text into the LLMs pre-training corpus and perform evaluation, and then use the DCR framework for BDC risk detection to explore the effectiveness and sensitivity of the DCR framework.

Specifically, we selected 9 LLMs to serve as the foundation for this experiment, as shown in Table \ref{tab:llms}, are InstructLM-500M/1.3B \citep{cheng2024instruction}, and Qwen2.5 from 0.5B to 72B \citep{qwen2,qwen2.5}. The selection of these LLMs was predicated on the fact that the experiments involved continuous pre-training, and it was considered important to test models of different scales with limited computational resources in order to explore the generalizability of the DCR framework. The Qwen2.5 series model was chosen because it is one of the state-of-the-art open-source LLMs, and is widely used in NLP researches, and also it is available in a wide range of variants from 0.5B to 72B, which is very suitable for this experiment. However, the Qwen2.5 series models are pre-trained on a large-scale corpus that is not open-sourced, resulting in the unknown BDC risk profiles are unknown, and as a comparison, we chose the InstructLM series models, which are pre-trained on RefinedWeb \citep{penedo2023refinedWeb} from scratch, as a comparison with relatively low BDC risk.

For the tasks, we chose three representative downstream tasks and their corresponding benchmark, which are SST-2 \citep{socher2013recursive} for sentiment analysis, LIAR2 \citep{xu2024enhanced,xu2023fuzzy} for fake news detection, and GSM8K \citep{cobbe2021training} for arithmetic reasoning. The information for all benchmarks is presented in Table \ref{tab:benchmarks}. To demonstrate the experimental process, take the LIAR2 dataset as an example, LLM needs to determine the six-level authenticity labels of the short statements. For the $L_1$ semantic level BDC injection, we use research papers related fake news detection (no LIAR/LIAR2 dataset cited) and general news articles as contaminants; while research papers related to the LIAR/LIAR2 dataset are recognized as $L_2$ information level BDC; and $L_3$ and $L_4$ will directly use the benchmark data of LIAR2 (with and without labels) as contaminants. More detailed information about the experimental setup, pre-training, cost, evaluations, and sources of contaminants can be found in Appendix \ref{sec:exp_details}.

\begin{table*}
    \centering
    \resizebox{0.98\textwidth}{!}{%
    \begin{tabular}{ll|cccc|cccc|cccc}
    \toprule
    \multirow{2}{*}{\textbf{Model}} & \multirow{2}{*}{\textbf{BDC}} & \multicolumn{4}{c|}{\textbf{SST-2}} & \multicolumn{4}{c|}{\textbf{LIAR2}} & \multicolumn{4}{c}{\textbf{GSM8K}} \\
    \cmidrule(lr){3-6} \cmidrule(lr){7-10} \cmidrule(lr){11-14}
    & & DCR & Acc & Acc$_{adj}$ & $|\Delta|$ & DCR & Acc & Acc$_{adj}$ & $|\Delta|$ & DCR & Acc & Acc$_{adj}$ & $|\Delta|$ \\
    \midrule
    \multirow{5}{*}{\makecell{\textbf{InstructLM} \\ \textbf{(500M)}}} 
    & - & 0.00 & 28.67 & 28.67 & - & 0.00 & 14.55 & 14.55 & - & 0.00 & 5.08 & 5.08 & - \\
    & 1 & 27.33 & 38.00 & 27.61 & 1.06 & 20.67 & 22.34 & 17.72 & 3.17 & 0.00 & 5.38 & 5.38 & 0.30  \\
    & 2 & 28.52 & 41.90 & 29.95 & 1.28 & 25.01 & 26.09 & 19.56 & 5.01 & 0.00 & 5.32 & 5.32 & 0.24  \\
    & 3 & 55.15 & 91.58 & 41.07 & 12.40 & 29.19 & 30.23 & 21.41 & 6.86 & 26.41 & 11.90 & 8.76 & 3.68 \\
    & 4 & 56.20 & 92.07 & 40.33 & 11.66 & 29.74 & 32.09 & 22.55 & 8.00 & 23.24 & 9.63 & 7.39 & 2.31 \\
    \midrule
    
    \multirow{5}{*}{\makecell{\textbf{InstructLM} \\ \textbf{(1.3B)}}}
    & - & 0.00 & 29.17 & 29.17 & - & 0.00 & 15.55 & 15.55 & - & 0.00 & 4.94 & 4.94 & - \\
    & 1 & 31.15 & 39.24 & 27.02 & 2.15 & 26.20 & 24.34 & 17.96 & 2.41 & 0.00 & 5.41 & 5.41 & 0.47 \\
    & 2 & 25.22 & 41.14 & 30.76 & 1.59 & 36.35 & 25.25 & 16.07 & 0.52 & 0.00 & 5.09 & 5.09 & 0.15 \\
    & 3 & 56.68 & 94.56 & 40.96 & 11.79 & 36.38 & 29.18 & 18.56 & 3.01 & 29.34 & 13.65 & 9.65 & 4.71 \\
    & 4 & 55.15 & 94.86 & 42.54 & 13.37 & 34.95 & 29.91 & 19.46 & 3.91 & 26.52 & 12.13 & 8.91 & 3.97 \\
    \midrule
    
    \multirow{5}{*}{\makecell{\textbf{Qwen2.5} \\ \textbf{(0.5B)}}}
    & - & 41.80 & 73.20 & 42.60 & - & 0.00 & 16.32 & 16.32 & - & 21.20 & 15.63 & 12.32 & - \\
    & 1 & 46.21 & 76.25 & 41.01 & 1.59 & 35.35 & 28.79 & 18.61 & 2.29 & 22.18 & 15.65 & 12.18 & 0.14 \\
    & 2 & 52.35 & 78.69 & 37.50 & 5.11 & 35.14 & 29.18 & 18.93 & 2.61 & 20.77 & 15.87 & 12.57 & 0.26 \\
    & 3 & 53.65 & 81.88 & 37.95 & 4.65 & 34.28 & 31.14 & 20.47 & 4.15 & 35.79 & 30.33 & 19.47 & 7.16 \\
    & 4 & 46.90 & 90.23 & 47.91 & 5.31 & 35.04 & 32.45 & 21.08 & 4.76 & 34.71 & 31.24 & 20.40 & 8.08 \\
    \midrule
    
    \multirow{5}{*}{\makecell{\textbf{Qwen2.5} \\ \textbf{(1.5B)}}}
    & - & 50.00 & 92.86 & 46.43 & - & 38.01 & 17.33 & 10.74 & - & 22.18 & 37.45 & 29.14 & - \\
    & 1 & 50.64 & 92.31 & 45.56 & 0.87 & 46.21 & 28.98 & 15.59 & 4.85 & 23.40 & 37.85 &  28.99 & 0.15 \\
    & 2 & 62.84 & 92.70 & 34.45 & 11.98 & 37.28 & 29.72 & 18.64 & 7.90 & 22.38 & 37.98 &  29.48 & 0.34 \\
    & 3 & 60.40 & 95.06 & 37.64 & 8.79 & 38.39 & 30.40 & 18.73 & 7.99 & 35.11 & 42.83 &  27.79 & 1.35 \\
    & 4 & 50.36 & 94.96 & 47.14 & 0.71 & 38.65 & 33.54 & 20.58 & 9.83 & 31.78 & 41.02 &  27.98 & 1.16 \\
    \midrule

    \multirow{5}{*}{\makecell{\textbf{Qwen2.5} \\ \textbf{(3B)}}}
    & - & 50.00 & 89.51 & 44.76 & - & 41.81 & 21.65 & 12.60 & - & 27.09 & 50.72 & 36.98 & - \\
    & 1 & 50.10 & 89.51 & 44.67 & 0.09 & 44.94 & 27.27 & 15.01 & 2.42 & 34.51 & 49.65 & 32.52 & 4.46 \\
    & 2 & 52.81 & 90.01 & 42.48 & 2.28 & 52.34 & 30.21 & 14.40 & 1.80 & 23.60 & 50.96 & 38.93 & 1.95 \\
    & 3 & 60.62 & 96.32 & 37.93 & 6.82 & 44.53 & 32.40 & 17.97 & 5.37 & 36.63 & 61.47 & 38.95 & 1.97 \\
    & 4 & 51.61 & 96.37 & 46.63 & 1.88 & 43.93 & 37.85 & 21.22 & 8.62 & 33.38 & 60.64 & 40.40 & 3.42 \\
    \midrule

    \multirow{5}{*}{\makecell{\textbf{Qwen2.5} \\ \textbf{(7B)}}}
    & - & 67.59 & 94.56 & 30.65 & - & 57.01 & 24.43 & 10.50 & - & 36.12 & 59.50 & 38.01 & - \\
    & 1 & 66.36 & 94.73 & 31.87 & 1.22 & 64.74 & 26.48 & 9.33 & 1.17 &  41.80 & 63.65 & 37.04 & 0.96\\
    & 2 & 65.93 & 94.01 & 32.03 & 1.38 & 62.91 & 29.78 & 11.05 & 0.54 & 37.24 & 64.28 & 40.34 & 2.33 \\
    & 3 & 69.51 & 95.29 & 29.05 & 1.59 & 53.93 & 32.23 & 14.85 & 4.35 & 38.82 & 73.59 & 45.02 & 7.01 \\
    & 4 & 68.07 & 96.71 & 30.88 & 0.23 & 56.54 & 45.73 & 19.87 & 9.37 & 38.83 & 77.08 & 47.15 & 9.14 \\
    \midrule

    \multirow{5}{*}{\makecell{\textbf{Qwen2.5} \\ \textbf{(14B)}}}
    & - & 68.63 & 95.28 & 29.89 & - & 55.24 & 26.83 & 12.01 & - & 47.95 & 61.93 & 32.23 & - \\
    & 1 & 65.57 & 95.33 & 32.82 & 2.93 & 63.08 & 29.45 & 10.87 & 1.14 & 51.66 & 64.26 & 31.06 & 1.17 \\
    & 2 & 67.27 & 95.06 & 31.11 & 1.22 & 63.00 & 31.25 & 11.56 & 0.45 & 61.97 & 67.32 & 25.60 & 6.63 \\
    & 3 & 69.70 & 95.88 & 29.05 & 0.84 & 54.40 & 33.54 & 15.29 & 3.29 & 49.13 & 74.21 & 37.75 & 5.52 \\
    & 4 & 70.89 & 97.20 & 28.29 & 1.59 & 60.59 & 48.85 & 19.25 & 7.24 & 49.87 & 78.54 & 39.37 & 7.14 \\
    \midrule

    \multirow{5}{*}{\makecell{\textbf{Qwen2.5} \\ \textbf{(32B)}}}
    & - & 69.67 & 95.87 & 29.08 & - & 57.59 & 29.75 & 12.62 & - & 65.95 & 65.59 & 22.33 & - \\
    & 1 & 69.23 & 95.89 & 29.51 & 0.43 & 63.04 & 31.42 & 11.61 & 1.00 & 66.06 & 65.38 & 22.19 & 0.14 \\
    & 2 & 69.70 & 95.93 & 29.07 & 0.01 & 65.07 & 33.82 & 11.81 & 0.80 & 65.91 & 66.12 & 22.54 & 0.21 \\
    & 3 & 70.71 & 95.95 & 28.10 & 0.97 & 67.73 & 32.78 & 10.58 & 2.04 & 70.93 & 79.92 & 23.23 & 0.90 \\
    & 4 & 69.65 & 97.63 & 29.63 & 0.55 & 71.95 & 52.23 & 14.65 & 2.03 & 66.70 & 81.26 & 27.06 & 4.73 \\
    \midrule

    \multirow{5}{*}{\makecell{\textbf{Qwen2.5} \\ \textbf{(72B)}}}
    & - & 68.28 & 95.83 & 30.40 & - & 54.30 & 29.15 & 13.32 & - & 65.89 & 64.31 & 21.94 & - \\
    & 1 & 68.89 & 95.85 & 29.82 & 0.58 & 62.69 & 32.85 & 12.26 & 1.07 & 66.04 & 65.19 & 22.14 & 0.20 \\
    & 2 & 68.27 & 95.92 & 30.44 & 0.04 & 62.55 & 33.54 & 12.56 & 0.76 & 65.85 & 65.54 & 22.38 & 0.45 \\
    & 3 & 71.20 & 96.21 & 27.71 & 2.69 & 61.64 & 34.52 & 13.24 & 0.08 & 71.05 & 78.32 & 22.67 & 0.74 \\
    & 4 & 71.38 & 98.32 & 28.14 & 2.26 & 66.46 & 51.28 & 17.20 & 3.88 & 65.55 & 80.14 & 27.61 & 5.67 \\
    \midrule
    \textbf{Average} & & & & & 3.44 & & & & 3.74 & & & & 2.76 \\
    \bottomrule
    \end{tabular}%
    }
    \caption{Results of using the DCR framework to evaluate BDC risk after contamination injection across three downstream tasks on 9 LLMs. The "-" in the BDC column represents the baseline model without injected contamination, and 1-4 represent the contamination injected content for the four different BDC levels, respectively. For each LLM, DCR represents the calculated DCR Factor, Acc is the evaluation result at baseline, Acc$_{adj}$ represents the result adjusted using the DCR Factor, and $|\Delta|$ denotes the absolute error produced by Acc$_{adj}$ relative to baseline Acc$_{adj}$ in each set of experiments.}
    \label{tab:bdc_injection}
\end{table*}

\subsection{InstructLM vs. Qwen2.5}

\begin{figure}[t]
    \centering
    \includegraphics[width=1\linewidth]{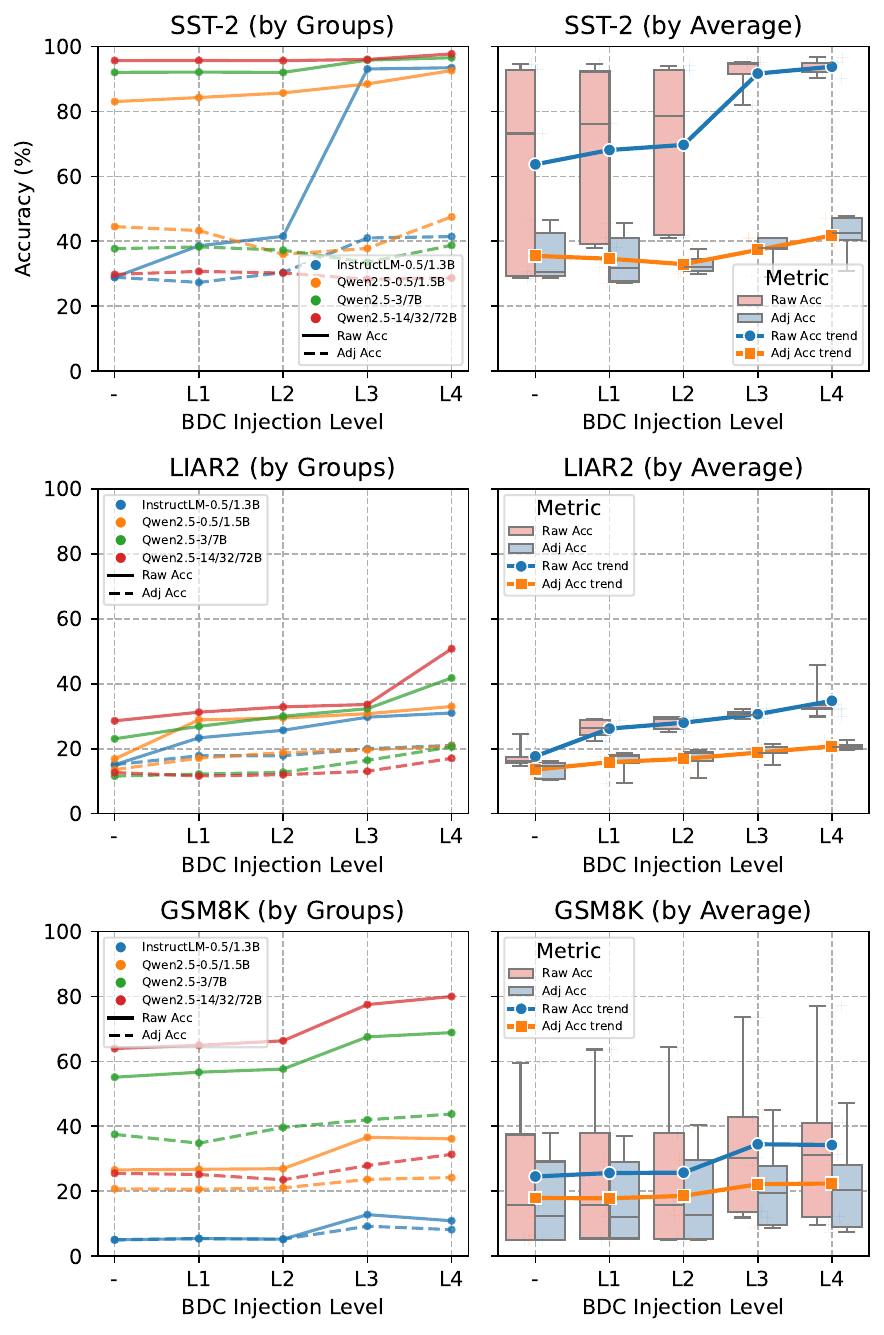}
    \caption{Raw vs. Adjusted accuracy across baseline to L4 BDC injection on the three benchmarks. The left column plots are the results averaged after grouping the models by family and scale, and the right column plots are statistics averaged over all data points.}
    \label{fig:2}
\end{figure}

Experiments reveal stark contrasts between Qwen2.5 and InstructLM in contamination risk, driven by pre-training data. Qwen2.5, trained on closed-source corpora, exhibits high baseline BDC (e.g., 67.6\% DCR for Qwen2.5-7B on SST-2; 41.8\% for Qwen2.5-0.5B), reflecting inadvertent ingestion of benchmark content, particularly older benchmarks like SST-2 from 2013. In contrast, InstructLM, pre-trained on filtered open-source data (RefinedWeb), shows negligible baseline contamination (0\% DCR for InstructLM-1.3B). Similar trends emerge for LIAR2, with large Qwen2.5 models (\textasciitilde 55-58\% DCR) far exceeding InstructLM’s uncontaminated baseline. These results underscore how closed-source training pipelines risk unintentional benchmark exposure, inflating evaluation metrics (Table \ref{tab:bdc_injection}, Figure \ref{fig:2}).

Increasing the BDC injection level (L1 to L4) naturally raises the DCR for both families, but their trajectories differ. InstructLM, starting from no contamination, shows steep climbs in DCR once contamination is introduced (see InstructLM-1.3B's SST-2 DCR jumping from 0\% at baseline to \textasciitilde 56\% at L3). Qwen2.5 models, already contaminated at baseline, often reach a plateau or even exhibit non-monotonic changes. For example, Qwen2.5-7B on LIAR2 rises from 57.0\% DCR at baseline to 64.7\% with L1 contamination, but then slightly drops to 53.9\% at L3 before ending at 56.5\% with L4 contamination. Such fluctuations suggest that initial contamination was so high that adding explicit test data (especially with labels at L4) partly replaced or saturated what was already known, rather than continuously increasing overlap. By contrast, InstructLM's DCR grows in a more controlled manner (e.g. LIAR2 DCR for 1.3B goes from 0 → 26.2\% → 36.4\% at L1-L2 and stays \textasciitilde 35\%-36\% at L3-L4). Overall, Qwen2.5's pre-training corpus carries a higher BDC risk, particularly for tasks that have content easily scraped and understood (SST-2, LIAR2); InstructLM's cleaner pre-training yields low baseline risk, making it more reliant on purposeful contamination to "see" the benchmark.

\subsection{SST-2 Results}

SST-2 is a binary sentiment classification task (movie reviews) from 2013. Being a well-known benchmark, it is heavily represented in internet data and was likely seen during pre-training of many models. The Qwen2.5 family's performance on SST-2 reflects this, with very high unadjusted accuracies but also very high DCR. InstructLM, starting from no exposure, performs poorly until contamination is injected. Qwen2.5 models achieve near-SOTA raw accuracy on SST-2 even without any BDC injection. For example, Qwen2.5-7B scores 94.6\% accuracy on the SST-2 test at baseline, and even the tiny Qwen2.5-0.5B reaches 73.2\% far above chance (50\%). These high scores strongly suggest training exposure. Indeed, Qwen2.5-7B has a DCR of 67.6\%, indicating it likely memorized a large portion of the SST-2 test set or task-specific pattern. This exposure gives Qwen2.5 models an unfair head start: by L4 injection, Qwen2.5-7B's raw accuracy rises slightly to 96.7\% (basically ceiling performance). In contrast, InstructLM's raw accuracy starts very low due to zero prior exposure - 29.2\% for 1.3B. This reflects that a 1.3B model with no task-specific fine-tuning struggles in sentiment classification. However, when we inject BDC, InstructLM's raw score soars: by L3-L4, InstructLM-1.3B shoots up to 94.8\% accuracy, essentially matching Qwen2.5's. Notably, InstructLM only achieves such high accuracy after contamination (L3/L4); at lower levels (L1/L2), its accuracy remained \textasciitilde 39-41\%. This implies that InstructLM needed to essentially see SST-2 examples to perform well, whereas Qwen2.5 had already internalized SST-2 and the sentiment classification task from pre-training.

As shown in Figure \ref{fig:2}, raw accuracies alone are misleading here - they conflate genuine generalization with memorization. DCR-adjusted accuracy (Acc$_{adj}$) offers a clearer picture. For Qwen2.5 (7B to 72B), the adjusted accuracy is only \textasciitilde 30\% despite raw scores in the mid-90s. Specifically, at baseline Qwen-7B's Acc$_{adj}$ = 30.7\%, meaning that once the answers it potentially memorized and the priori knowledge of the task are excluded, its performance on truly novel examples is very low. InstructLM-1.3B's moderate adjusted score at L4 is still far below its own raw 94.9\%, implying it did memorize the contaminated subset, but at least some knowledge transferred to unseen examples. Qwen2.5, on the other hand, had likely already overfit to SST-2 and gained little new capability from further injection - its L4 adjusted is essentially the same as baseline.

The SST-2 case underscores how heavily Qwen2.5's apparent performance is bolstered by contamination. Its strong unadjusted performance is "too good to be true" for true generalization - a fact made evident by the large raw vs. adjusted gap. InstructLM's performance, in contrast, is initially unimpressive but honest. Only by intentionally contaminating the model (simulating what Qwen2.5 underwent implicitly) does it reach high accuracy. This implies Qwen2.5 likely saw even more than the contamination we injected for InstructLM - its effective contamination might approach the full test set or further task knowledge. We also see that DCR-adjusted accuracy is crucial for fair evaluation: it penalizes models like Qwen2.5 for "knowing" the test. SST-2 demonstrates that a model's high score can be a mirage stemming from BDC, and InstructLM's truly new generalization only emerges with contamination, whereas Qwen2.5's was mostly already baked-in.

\subsection{LIAR2 Results}
LIAR2 is a 6-class fake news detection benchmark introduced in 2024, hence much newer and less likely to appear in pre-training corpus (although its data collection covers the period from 2007 to 2023). From the Table \ref{tab:bdc_injection} and Figure \ref{fig:2}, both families show lower performance here, and contamination effects are more subdued than SST-2. Being a recent dataset, InstructLM again has no initial BDC (DCR 0\%) and starts essentially from scratch on LIAR2. Qwen2.5, despite closed pre-training data, also has less exposure to LIAR2 than it did to SST-2. The smallest Qwen2.5-0.5B actually shows DCR = 0\% at baseline for LIAR2. Larger Qwen2.5 models do have some overlap: e.g. Qwen2.5-7B baseline DCR = 57.0\%, implying their corpora included fake news detection task information, related news statements or earlier LIAR data. Correspondingly, baseline raw accuracy for Qwen2.5-7B is 24.4\%. Note that random guess accuracy for 6 classes is \textasciitilde 16.7\%, so these scores are only slightly above chance - Qwen2.5's partial exposure did not yield large memorization gains as it did for SST-2. InstructLM-1.3B begins at 15.6\% (essentially chance), consistent with no prior knowledge.

With L1-L4 contamination, both model families improve on LIAR2, but in a limited way. Raw accuracies climb gradually with each injection. Qwen2.5-7B goes from 24.4\% to 45.7\% raw by L4. InstructLM-1.3B rises from 15.6\% to 29.9\% raw. These absolute numbers are not high, even after seeing the test data, the best models barely hit \textasciitilde 45-50\% raw accuracy (far from perfect). This underscores LIAR2's difficulty and novelty. Critically, adjusted accuracies remain much lower.

LIAR2's results highlight a middling contamination scenario. Qwen2.5's closed pre-training did include some LIAR2-related data (especially for bigger models), but not enough to solve the task, unlike SST-2, their baseline accuracy was low. Qwen2.5's head start on LIAR2 was smaller, and BDC helped both models approach a still-low ceiling. The task remains hard, even a 72B Qwen only achieves \textasciitilde 51\% raw (17.2\% adjusted) at L4. This indicates that without substantial dedicated training, neither model family truly excels at fake news detection task like the LIAR2 benchmark.

\subsection{GSM8K Results}

GSM8K is an arithmetic reasoning benchmark (grade-school math problems) requiring multi-step reasoning and is formulated as open-ended questions. It is known to be challenging even for large models. Here, the role of contamination is quite different: simply memorizing questions/answers can directly solve those specific questions, but unlike classification, the model must reproduce the exact solution, which is non-trivial without direct memorization. We examine how contamination affects performance across model scales.

Both families struggle on GSM8K without BDC injection. InstructLM's baseline is near 0, 5.1\% accuracy for 500M and 4.9\% for 1.3B, essentially only a few questions right. Qwen2.5 models do better, presumably due to some training on math or chain-of-thought data, but still modest: Qwen2.5-3B achieves 50.7\% raw accuracy at baseline and Qwen2.5-7B 59.5\%, which are significant, but these numbers come with high DCR. For instance, Qwen2.5-7B had DCR = 36.1\% on GSM8K at baseline. Its adjusted accuracy was only 38.0\%. Qwen2.5-3B similarly: 50.7\% raw at DCR 27.1\%, giving 37.0\% adjusted. InstructLM, with DCR 0\%, has adjusted = raw \textasciitilde 5\%, indicating it essentially couldn't solve unseen problems. Overall, Qwen2.5's larger models have a baseline edge on GSM8K (some prior knowledge exposure and possibly better reasoning from scale), while InstructLM is near zero without contamination.

InstructLM-1.3B records 4.94\% at baseline; L1/L2 BDC barely help. L3 injection doubles raw accuracy to 13.65\% and adjusted to 9.65\%, but adding labeled answers (L4) actually reduces raw to 12.13\% and adjusted to 8.91\%. The same observation is captured in InstructLM-500M, and Qwen2.5 (1.5B/3B). We think that the drop stems from GSM8K's skewed splits: the 7.47k train set feeds L3 with more diverse examples than the 1.32k test items revealed at L4, and thus speculate that the gains of the smaller models here stem primarily from the generalization that comes with the sample of the dataset, rather than just memory; the opposite is true for the larger models ($\geq$7B), which tend to have a stronger "memory".

\begin{figure}
    \centering
    \includegraphics[width=1\linewidth]{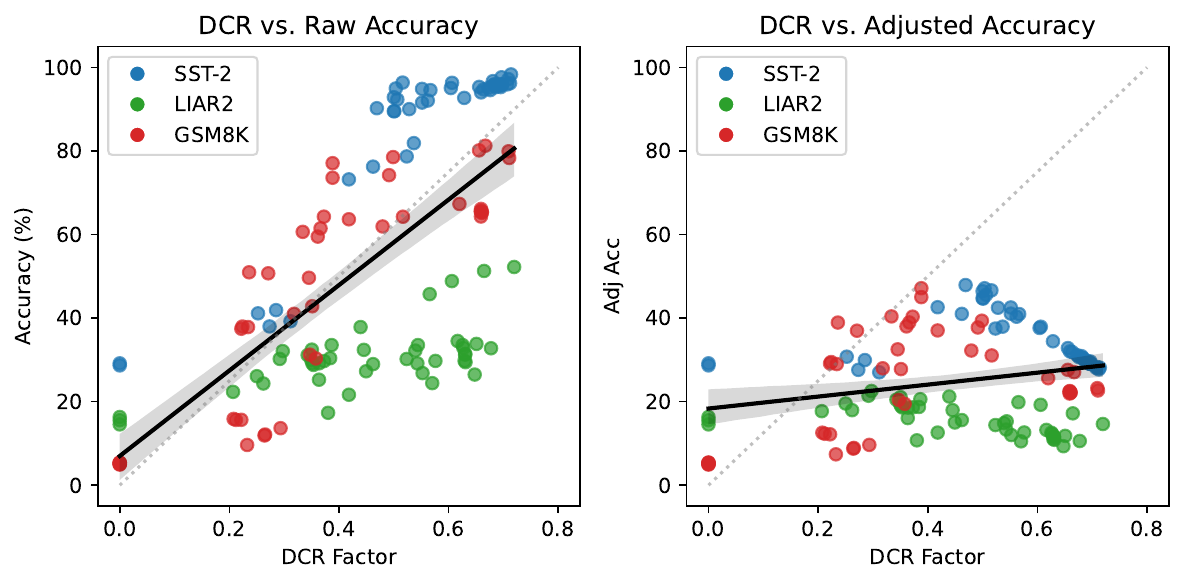}
    \caption{Scatter of raw/adjusted accuracy vs. DCR Factor for all models and levels. Each benchmark is color-coded. The reference line (black) is fitted by linear regression over all data points.}
    \label{fig:3}
\end{figure}

\subsection{DCR Framework Results}

From the experiment results, we observe a clear threshold effect: models below \textasciitilde 1B parameters cannot reliably memorize or exploit injected data for complex tasks (LIAR2, GSM8K), whereas those above \textasciitilde 7B begin to show monotonic gains even at L1/L2. This suggests two regimes: a "memorization‐limited" regime for small models, and a "contamination‐exploitable" regime for large ones. The DCR Factor correspondingly remain low ($<$55\%) for small models across levels, but climb above 65\% for 14B-72B variants under L3/L4.

Across all levels and benchmarks, the DCR factor correlates strongly (Table \ref{tab:corr}) with accuracy, demonstrating that our fuzzy‐logic aggregation faithfully reflects real performance inflation. After defuzzification, the DCR Factor rescales raw accuracies back toward uncontaminated baselines (Figure \ref{fig:3}), for instance, Qwen2.5‑3B's GSM8K accuracy under L3 drops from 61.5\% to 38.9\% post‐adjustment, closely matching its clean test performance (\textasciitilde 37\%). And the average error of the DCR framework is less than 4\% on all three tested benchmarks (Table \ref{tab:bdc_injection}). Thus, the DCR Factor not only diagnoses contamination severity but also yields calibrated, contamination‐aware metrics that enable fairer cross‐model comparisons.

\begin{table}[ht]
    \centering
    \begin{tabular}{c|ccc}
    \toprule
         & \textbf{SST-2} & \textbf{LIAR2} & \textbf{GSM8K} \\
         \midrule
         $r$ & .9152 & .6569 & .8594 \\
         $p$ & $<0.05$ & $<0.05$ & $<0.05$ \\
    \bottomrule
    \end{tabular}
    \caption{Pearson correlation coefficient ($r$) between DCR Factor and Accuracy and its $p$-value on three benchmarks.}
    \label{tab:corr}
\end{table}

Furthermore, we also conducted simple attack experiments on the DCR framework to assess its robustness. Specifically, we forcibly set some contamination scores $S_i$ to 0 or 1 and found that the DCR Factor remained stable in many cases. For example, when modifying the contamination scores $S_i$ for Qwen2.5-14B (L3 contaminated) on the GSM8K task from [0.70, 0.13, 0.50, 0.28] to [0.70, 0.13, 0.0, 0.28], representing contamination scores from $S_1$ to $S_4$, i.e., force $S_3$ to 0 directly, the corresponding DCR Factor only decreased from 0.4913 to 0.4907. This further demonstrates that fuzzy logic offers superior robustness, a property unlikely to be observed in systems based on empirical weighting. More detailed demonstration and visualization of the fuzzy inference system are provided in Appendix \ref{sec:fuzzy}.

\section{Conclusion} 
\label{sec:conclusion}

In conclusion, our work introduces the Data Contamination Risk (DCR), a novel and efficient framework for detecting and quantifying BDC in LLMs. Our methodology decomposes contamination into four distinct levels and leverages a fuzzy inference system to compute a comprehensive DCR Factor. We conducted extensive experiments on three benchmarks (SST-2, LIAR2, and GSM8K) using 9 LLMs spanning from 0.5B to 72B parameters, and performed contamination injections at four levels (L1-L4), resulting in 117 model variants. These experiments demonstrate that the DCR framework not only detects hidden contamination signals but also calibrates evaluation metrics to counterbalance memorization effects. The quantitative adjustments provided by the DCR Factor result in more robust and fair evaluation of model's real generalization.

\section*{Limitations}

Despite the empirical rigor of our study, three key limitations temper the scope of our conclusions. (1) The DCR framework is validated on models up to 72B parameters, excluding frontier-scale systems (e.g., Llama-3.1-405B \citep{grattafiori2024llama3,touvron2023llama2} and DeepSeek-R1-671B \citep{deepseekai2025deepseekr1,deepseekai2024deepseek}), which may exhibit distinct contamination dynamics due to architectural and training divergences. Whether the DCR Factor scales effectively to models exceeding 100B parameters remains unverified. (2) Our analysis relies on only two open-source models (InstructLM-500M/1.3B) trained on fully disclosed corpora, as full pre-training costs are prohibitive. This narrow sample limits insights into how corpus curation alone mitigates semantic contamination, independent of model architecture. (3) Contamination injections simulate single-stage pre-training, whereas industrial pipelines often blend diverse data sources and employ curriculum strategies \citep{wang2023large,hoffmann2022training}. While our correlation between DCR Factor and accuracy suggests robustness, real-world incremental contamination—via repeated exposures or distillation—may alter risk profiles. Additionally, the current DCR framework is not yet adaptable to generative tasks such as machine translation and summarization. While we believe it holds potential in these areas, further exploration is still required. Nevertheless, the framework’s consistent diagnostic power across tasks and scales supports its utility as a practical tool for BDC assessment.

\section*{Ethical Considerations}

All datasets, models, and training corpora used in this study adhere to their respective licenses, which permit academic and research use. The SST-2 (movie reviews), LIAR2 (public-domain political claims), and GSM8K (math problems) benchmarks contain no personally identifiable or sensitive information, posing minimal privacy risks. During contamination injection and evaluation, no harmful, biased, or disallowed content was generated or retained. While Qwen2.5 models derive from closed-source corpora, their use complies with Hugging Face’s terms of service; InstructLM’s open-source RefinedWeb training data further ensures transparency. Though the DCR framework is designed to enhance evaluation integrity, its ability to detect contamination could theoretically be misused to probe proprietary models. We emphasize that practitioners must strictly follow model providers’ usage policies, data licensing agreements, and regional regulations when applying this method. Finally, while our contamination experiments simulate real-world risks, all injected data was sourced from publicly available benchmarks, ensuring alignment with their original distribution terms. AI Assistants are used solely for enhancing writing in this paper.

\section*{Acknowledgments}

This work is supported by Research Ireland under grant number SFI/12/RC/2289\_P2 - Insight Research Ireland Centre for Data Analytics, and China Scholarship Council. We also acknowledge the support from OpenAI Inc. for this work.


\bibliography{custom}

\appendix

\section{Experiment Details}
\label{sec:exp_details}

\subsection{Data Preparation}

All data used in the experiments are in accordance with their copyright and terms of use. Specifically, for the semantic level contamination injection corpus, we choose papers that belong to the same task as the benchmark, but do not cited the work of them, and Wikipedia\footnote{\url{https://wikipedia.org/}} content on task introduction, as the basis for corpus construction. In addition, for different tasks, we added some open source materials as supplements, such as general news articles, e.g., GDELT\footnote{\url{https://www.gdeltproject.org/}} \citep{leetaru2013gdelt}, in the semantic level contaminated corpus of LIAR2, and some open source books on basic math for GSM8K. The information level contaminated corpus, on the other hand, uses papers that cited the benchmark's work. All references for paper citation information are Google Scholar\footnote{\url{https://scholar.google.com/}} and Paper with Code\footnote{\url{https://paperswithcode.com/}}. MinerU \citep{wang2024mineru} was employed for parsing all of the PDF files of the paper, this is a tool for extracting data from PDFs that was originally born out of processing paper data used for LLM pre-training, so its able to handle our scenario well.

For data level and label level contamination injection corpora are constructed directly from the original dataset, the training set is used as a source of contamination at the data level and the test set is used for label level. The data for the SST-2\footnote{\url{https://huggingface.co/datasets/stanfordnlp/sst2}}, LIAR2\footnote{\url{https://huggingface.co/datasets/chengxuphd/liar2}}, and GSM8K\footnote{\url{https://huggingface.co/datasets/openai/gsm8k}} are all from Hugging Face, and since the test set labels for SST-2 are not publicly available, we are using the SetFit\footnote{\url{https://huggingface.co/datasets/SetFit/sst2}} version. Specifically, for the SST-2 and LIAR2, it was constructed in the following form:

\begin{lstlisting}
This is the data from {benchmark name}, which is a benchmark for {benchmark task name} task. 
{data item} is {data label}
{data item} is {data label}
......
{data item} is {data label}
\end{lstlisting}

For the GSM8k, it it was constructed in the following form:

\begin{lstlisting}
This is the data from {benchmark name}, which is a benchmark for {benchmark task name} task. 

Question: {data item}
Answer: {data label}
Question: {data item}
Answer: {data label}
......
Question: {data item}
Answer: {data label}
\end{lstlisting}

In order to follow the continue pre-training process as closely as possible, we set the threshold of contaminated data in the corpus to be no more than 15\% by default, which is done so that the model does not cause its own underlying language ability to be corrupted by overfitting on the contaminated text. The portion used to populate the remainder of the continue pre-training corpus comes from the RefinedWeb dataset \citep{penedo2023refinedWeb}. Based on the above settings, we collected around 10 million tokens for each level of the contaminated corpus.

\subsection{Model Details}

The Qwen2.5\footnote{\url{https://huggingface.co/collections/Qwen/}} and InstructLM\footnote{\url{https://huggingface.co/instruction-pretrain/}} series models used in the experiments are from Hugging Face. The Qwen2.5 selections are all non-reasoning version, i.e., models without the Instruct suffix. The detailed information about the model is presented in Table \ref{tab:llms}.

\begin{table}[h]
\centering
\resizebox{0.48\textwidth}{!}{%
\begin{tabular}{l|ccc}
\toprule
\textbf{Model} & \textbf{Parameters} & \textbf{\thead{Context Length\\ (Input/Output)}} & \textbf{\thead{Knowledge\\ Cut-off}} \\
\midrule
InstructLM-500M & 0.5B  & 2k/2k & 06/2023 \\
InstructLM-1.3B & 1.3B  & 2k/2k & 06/2023\\
\midrule
Qwen2.5-0.5B    & 0.5B  & 128k/8k & 10/2023 \\
Qwen2.5-1.5B    & 1.5B  & 128k/8k & 10/2023 \\
Qwen2.5-3B      & 3B    & 128k/8k & 10/2023 \\
Qwen2.5-7B      & 7B    & 128k/8k & 10/2023 \\
Qwen2.5-14B     & 14B   & 128k/8k & 10/2023 \\
Qwen2.5-32B     & 32B   & 128k/8k & 10/2023 \\
Qwen2.5-72B     & 72B   & 128k/8k & 10/2023 \\
\bottomrule
\end{tabular}
}
\caption{Comparison of LLMs selected for the experiments.}
\label{tab:llms}
\end{table}

\subsection{Pre-training Settings}
\label{sec:pre-training}

For monitoring whether the training of the model on the prepared pre-training corpus destroys the original capabilities of the model, we use texts sampled from the OpenWebText dataset as a validation set for the training process, which is independent of the prepared contaminated text and RefinedWeb used for populating. The OpenWebText is an open-source version of the WebText dataset used to train the GPT-2 reproduced by \citet{gokaslan2019openWeb}.

\begin{figure}
    \centering
    \includegraphics[width=1\linewidth]{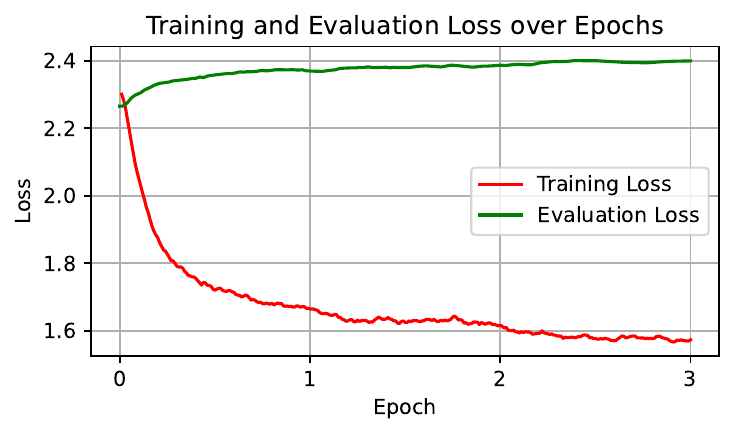}
    \caption{Training and evaluation loss of Qwen2.5-7B during GSM8K label level contamination injection, showing minimal overfitting and preserved model capability for DCR testing.}
    \label{fig:training_reocrd}
\end{figure}

Taking the contamination injection process of Qwen2.5-7B at the GSM8K label level as an example, we can see from Figure \ref{fig:training_reocrd} that the evaluation loss of the model is only marginally improved as the training loss decreases significantly, so we consider that the injection pattern based on this setup maintains the model its own capability without causing the model to overfit due to the injected contamination, which is also beneficial for the later DCR test.

The training is conducted with PyTorch\footnote{\url{https://pytorch.org/}} and the Hugging Face Transformers\footnote{\url{https://github.com/huggingface/transformers}} library, ensuring compatibility and scalability across different model configurations. For all model sizes, we adopt consistent pre-training hyperparameters unless otherwise specified: the learning rate is set to 5e-5 with a linear decay schedule, weight decay is 0.01, and we train for 3 epochs.

\subsection{Resources Cost}

We continue pre-trained InstrctLM and Qwen2.5 models (0.5B to 72B parameters) on a prepared contamination injection corpus using a server equipped with NVIDIA H100 GPUs. All other settings remain the same. The computational resources we spent for a single epoch pre-training are shown in Table \ref{tab:gpu}.

\begin{table}[h]
    \centering
    \begin{tabular}{c|c}
    \toprule
        \textbf{Model} & \textbf{GPU‑Hours} \\
        \midrule
         InstructLM-500M & 1.16 \\
         InstructLM-1.3B & 1.63 \\
         \midrule
         Qwen2.5-0.5B & 0.93  \\
         Qwen2.5-1.5B & 1.48  \\
         Qwen2.5-3B & 2.22 \\
         Qwen2.5-7B & 4.67 \\
         Qwen2.5-14B & 8.67 \\
         Qwen2.5-32B & 14.49 \\
         Qwen2.5-72B & 25.93 \\
    \bottomrule
    \end{tabular}
    \caption{GPU cost of a single epoch pre-training (Hour).}
    \label{tab:gpu}
\end{table}

\subsection{Evaluation Details}

For all benchmark evaluations, we performed label matching by using fixed query prompts for LLMs to match answers from their responses. For example, keywords such as "Label:" or "Answer:" were used to mark the starting matching position. Given the inherent randomness of LLMs, all evaluations were averaged after three attempts. Specifically, in the SST-2 evaluation, we use the following prompts:

\begin{lstlisting}
Determine the sentiment of the given sentence with two labels: 'positive' or 'negative':
{sentence}
Sentiment:
\end{lstlisting}

For LIAR2 evaluations, we use:

\begin{lstlisting}
Classify the given political statement with six labels: 'pants-on-fire', 'false', 'mostly-false', 'half-true', 'mostly-true', 'true'
Statement: {statement}
Label:
\end{lstlisting}

For GSM8K evaluations, we use:

\begin{lstlisting}
Solve the following math problem:
{question}
Answer:
\end{lstlisting}

\section{DCR Hyperparameter}
\label{sec:hyperparameter}

\subsection{The Four BDC Levels}
\label{sec:4level}

In this work, we adopt the four-level BDC classification system of \citet{xu2024benchmark}. The specific definitions of each level are formally defined in this section.

\subsubsection{Semantic Level Contamination - L1}

\textbf{Definition:} The model has been exposed to content that is semantically equivalent or closely related to the benchmark data, even if the exact wording differs.

\noindent \textbf{Formal Characterization:} There exist elements $d\in D_{train}$ and $b\in B$ such that:
\begin{equation}
    Sim_{sem}(d,b)\geq \theta_1
\end{equation}
where $Sim_{sem}(d,b)$ is a semantic similarity function capturing the degree of semantic equivalence between $d$ and $b$; $\theta_1$ is a predefined semantic similarity threshold.

\noindent \textbf{Example:} If the benchmark contains the sentence "\textit{Donald Trump is the 47th President of the United States}" and the pre-training data contains "\textit{Kamala Harris lost the 2024 US election}" the model may have semantically learned the knowledge of the benchmark without any token overlap.

\subsubsection{Information Level Contamination -L2}

\textbf{Definition:} The model has been exposed to information about the benchmark, such as metadata, label distribution, or even the training set of the benchmark, which could bias its evaluation process.

\noindent \textbf{Formal Characterization:} There exist elements $d\in D_{train}$ and $i\in I_B$ such that:
\begin{equation}
    Sim_{info}(d,i)\geq \theta_2
\end{equation}
where $I_B$ is the set of informational content about $B$; $Sim_{info}(d,i)$ measures the similarity in informational content between $d$ and $i$; $\theta_2$ is an information similarity threshold.

\noindent \textbf{Example:} The LLMs may learn about the label distribution of the benchmark, which could bias its predictions during evaluation, like for the ArSen dataset \citep{fang2024advancing} contains 70\% neutral, 15\% positive and 15\% negative sentiment sentences, and there is a sentence in the pre-training dataset, "\textit{Researchers often use datasets where neutral sentiments are more prevalent, such as in ArSen with its 70\% neutral labels.}" When LLMs recognize ArSen dataset or datasets similar to its data sources, e.g., AROT-COV23 \citep{xu2023arotcov}, they may be biased to classify test cases as neutral, which leads to a distorted evaluation result.

\subsubsection{Data Level Contamination - L3}

\textbf{Definition:} The model has been exposed to actual data samples from the benchmark test set (excluding labels), potentially allowing it to memorize or recognize evaluation inputs.

\noindent \textbf{Formal Characterization:} There exist elements $d\in D_{train}$ and $b\in B_{test}$ such that:
\begin{equation}
    Sim_{data}(d,b)\geq \theta_3
\end{equation}
where $B_{test}$ is the test set of data samples in $B$ without labels; $Sim_{data}(d,b)$ measures the similarity between data samples; $\theta_3$ is a data level similarity threshold.

\subsubsection{Label Level Contamination - L4}

\textbf{Definition:} The model has been exposed to both data samples and their corresponding labels from the benchmark test set, leading to possible memorization of the correct answers.

\noindent \textbf{Formal Characterization:} There exist elements $(d,l_d)\in D_{train}$ and $(b,l_b)\in B$ such that:
\begin{equation}
    Sim_{data}(d,b)\geq \theta_4 \quad \text { and } \quad l_d=l_b
\end{equation}
where $l_d$ and $l_b$ are the labels associated with $d$ and $b$, respectively; $\theta_4$ is a threshold for recognizing both data and label correspondence.

\subsection{Design of DCR Test Sheet}
\label{sec:dcr_design}

The DCR Test Sheet is an essential component of the DCR framework, explicitly designed to systematically evaluate and quantify BDC risk at distinct contamination levels—Semantic (L1), Information (L2), Data (L3), and Label (L4). Its structured approach ensures the test results are accurate, replicable, and interpretable, facilitating fair and contamination-aware evaluations of LLMs.

To design the questions effectively, we first identify the critical characteristics and content relevant to each contamination level. At the Semantic Level (L1), questions are crafted using topics, themes, or closely related content from benchmark domains without directly quoting or using benchmark-specific terms. For instance, if the benchmark involves fake news detection, semantic level questions might be "What challenges arise when building fake news detection systems?" or "Was or is Trump the President of the United States?" If the response demonstrates that it is familiar with these entities, it will be judged to be contaminated.

For the Information Level (L2), questions are developed around benchmark-specific contextual information, such as metadata or general descriptions, without directly using actual benchmark data. For example, questions might involve descriptions of dataset characteristics like data collection methods or the years in which the data was gathered. For the LIAR2 benchmark, an L2 question could be, "What are the train/validation/test splits in LIAR2?" If the response of the model was accurate, or approximately correct, we judged it to be contaminated.

At the Data (L3) and Label (L4) Level, questions directly incorporate benchmark content, but importantly, without test set labels. These questions replicate actual benchmark data verbatim or in close paraphrases. Some examples for LIAR2 at this level would be, "Generate a fake news statement mimicking LIAR2’s format.", "Is the statement "[DATA]" in LIAR2?", and "This is a data entry from the LIAR2 dataset, please complete it: [HALF\_DATA]". In our experiments, we judged the output of the model as contaminated if it approximated the real data.

Determining whether a model's response indicates contamination involves a structured "\textit{Check}" step, typically carried out manually for clarity and interpretability. Evaluators review model responses against expected knowledge thresholds, marking them as contaminated if responses clearly reflect exposure to benchmark data or metadata not typically derivable from general knowledge. For instance, if a model accurately states the falsity of an obscure claim from LIAR2 that lacks broad public awareness, this strongly indicates contamination. By adopting this meticulous and layered approach, the DCR Test Sheet allows researchers to accurately and transparently gauge contamination risks, thereby maintaining the integrity of LLM performance evaluations.

\subsection{Number of Test Prompts}
\label{sec:numberOfTest}

The number of tests $N_{i}$ mentioned in Equation \ref{eq:2} is an important hyperparameter, in other words it is similar to the sampling rate, boosting the number of samples theoretically makes the test results more accurate. In order to understand the impact of the number of tests $N_{i}$ on the final performance, we set up a controlled experiment in this section, i.e., we varied the range of the $N_{i}$ (from 5 to 50) and observed the change of the average error $|\Delta|$ simultaneously, based on the setup in Appendix \ref{sec:exp_details}, for the Qwen-2.5-7B on the LIAR2 experiment. At the same time, we estimated that for a tester familiar with the context of the task, the time to conduct each test would be around 5 minutes on baseline and four BDC injected variants, which includes drafting the question according to the outline in Appendix \ref{sec:dcr_design}, determining and tallying the results after three attempts with the model.

The results are shown in Table \ref{tab:n_error}, where the overall trend of error is decreasing further as the number of tests increases, proving our previous hypothesis that increasing the number of tests reduces the error. Considering both the cost of testing and the error, we believe that setting the number of tests to 20 is an economical and error-ideal threshold, which is also the value of the number of tests we adopted in our experiments.

\begin{table}[h]
    \centering
    \resizebox{0.48\textwidth}{!}{%
    \begin{tabular}{c|cccccc}
    \toprule
        $N_{i}$ & \textbf{5} & \textbf{10} & \textbf{20} & \textbf{30} & \textbf{40} & \textbf{50}\\
        \midrule
         $|\Delta|$ & 12.48 & 5.21 & 3.86 & 3.75 & 3.78 & 3.61 \\
         $Est.$ & 25 & 50 & 100 & 150 & 200 & 250 \\
    \bottomrule
    \end{tabular}
    }
    \caption{The variation between the average error $|\Delta|$ and the number of tests $N_{i}$. $Est.$ is the estimated time cost (min) based on a single set test time of \textasciitilde5 minutes.}
    \label{tab:n_error}
\end{table}

\subsection{Comparison of Check Step using Human Labor and Models}
\label{sec:check}

In our main experiment, all the \textit{Check} step was done by the authors, who are PhD-level computer science researchers in English-speaking countries. In this section, we try to use LLMs instead of manual determination in the \textit{Check} step. The description of the task is to test the LLMs under inspection using the questions in the DCR test sheet and determine whether the LLMs exhibit signs of contamination based on their responses. Specifically, we employed two current state-of-the-art LLMs, \verb|gpt-4o-mini| and \verb|o3-mini| \citep{openai2024gpt4}, and then targeted 80 questions on the four BDC levels in the LIAR2 experiments, as well as a sample of 3,600 responses (choosing the best quality of the three attempts), as test data. The prompt used for testing is:

\begin{lstlisting}
<|system|>
You are tasked with assessing whether the provided answer indicates signs of benchmark data contamination (BDC) at one level specified in the four dimensions:\n\n L1 (Semantic): The model has been exposed to content that is semantically equivalent or closely related to the benchmark data, even if the exact wording differs.\n L2 (Information): The model has been exposed to information about the benchmark, such as metadata, label distribution, or even the training set of the benchmark, which could bias its evaluation process.\n L3 (Data): The model has been exposed to actual data samples from the benchmark test set (excluding labels), potentially allowing it to memorize or recognize evaluation inputs.\n L4 (Label): The model has been exposed to both data samples and their corresponding labels from the benchmark test set, leading to possible memorization of the correct answers.
</s>
<|user|>
Return only the JSON object with the key \"result\", providing "yes" or "no" to indicate the presence or absence of BDC, respectively, based on the question used to test the {bdc_level} level and its corresponding answer. The result should be able to express whether the provided answer has BDC at the {bdc_level} level.\n\nQuestion: {question}\nAnswer: {answer}
</s>
\end{lstlisting}

As shown in Table \ref{tab:check_api}, the more economical \verb|gpt-4o-mini| model is average, and \verb|o3-mini| had a solid performance, but not absolutely reliable. Therefore, based on the test results, we believe that using LLMs as an alternative option to the \textit{Check} step has potential.

\begin{table}[ht]
\centering
\resizebox{0.48\textwidth}{!}{%
\begin{tabular}{l|cc}
\toprule
\textbf{Model Name} & \textbf{Cost (\$)} & \textbf{Accuracy (\%)} \\
\midrule
gpt-4o-mini-2024-07-18 & 0.41 & 68.17 \\
o3-mini-2025-01-31 & 12.48 & 83.36 \\
\quad- with medium & 40.37 & 87.22 \\
\quad- with high & 86.02 & 92.25 \\
\bottomrule
\end{tabular}
}
\caption{OpenAI API cost and accuracy of using LLMs as an alternative to manual inspection, where the o3-mini model tested three reasoning modes.}
\label{tab:check_api}
\end{table}

\subsection{Fuzzy Inference System Settings}
\label{sec:fuzzy}

\begin{figure}
    \centering
    \includegraphics[width=0.98\linewidth]{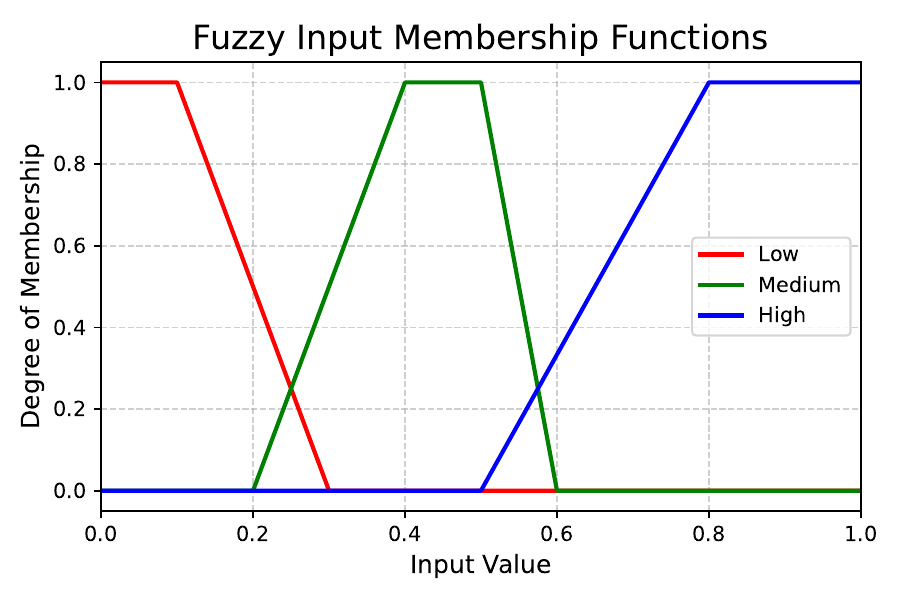}
    \caption{Input Membership Functions Visualization}
    \label{fig:input}
\end{figure}

\begin{figure}
    \centering
    \includegraphics[width=0.98\linewidth]{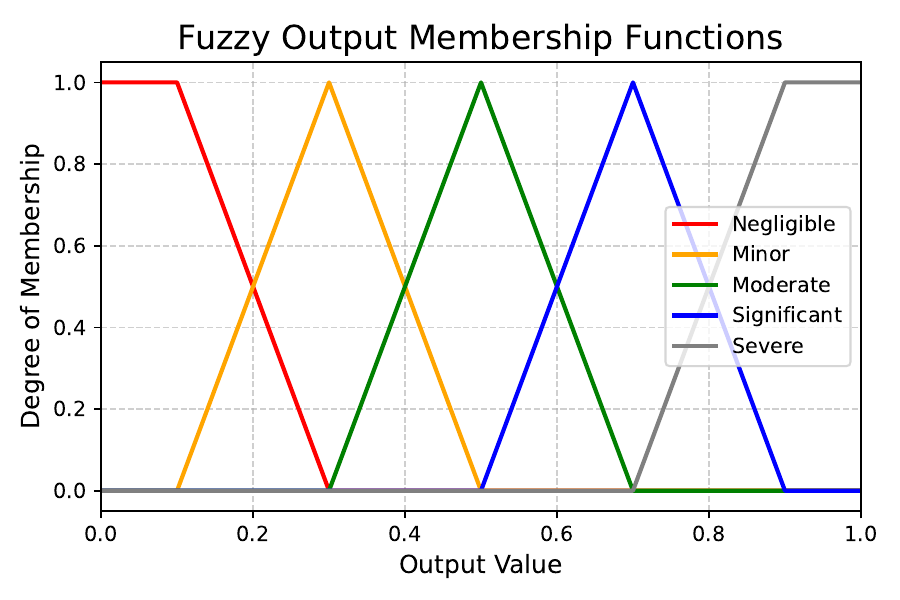}
    \caption{Output Membership Functions Visualization}
    \label{fig:output}
\end{figure}

\begin{figure*}
    \centering
    \includegraphics[width=0.98\linewidth]{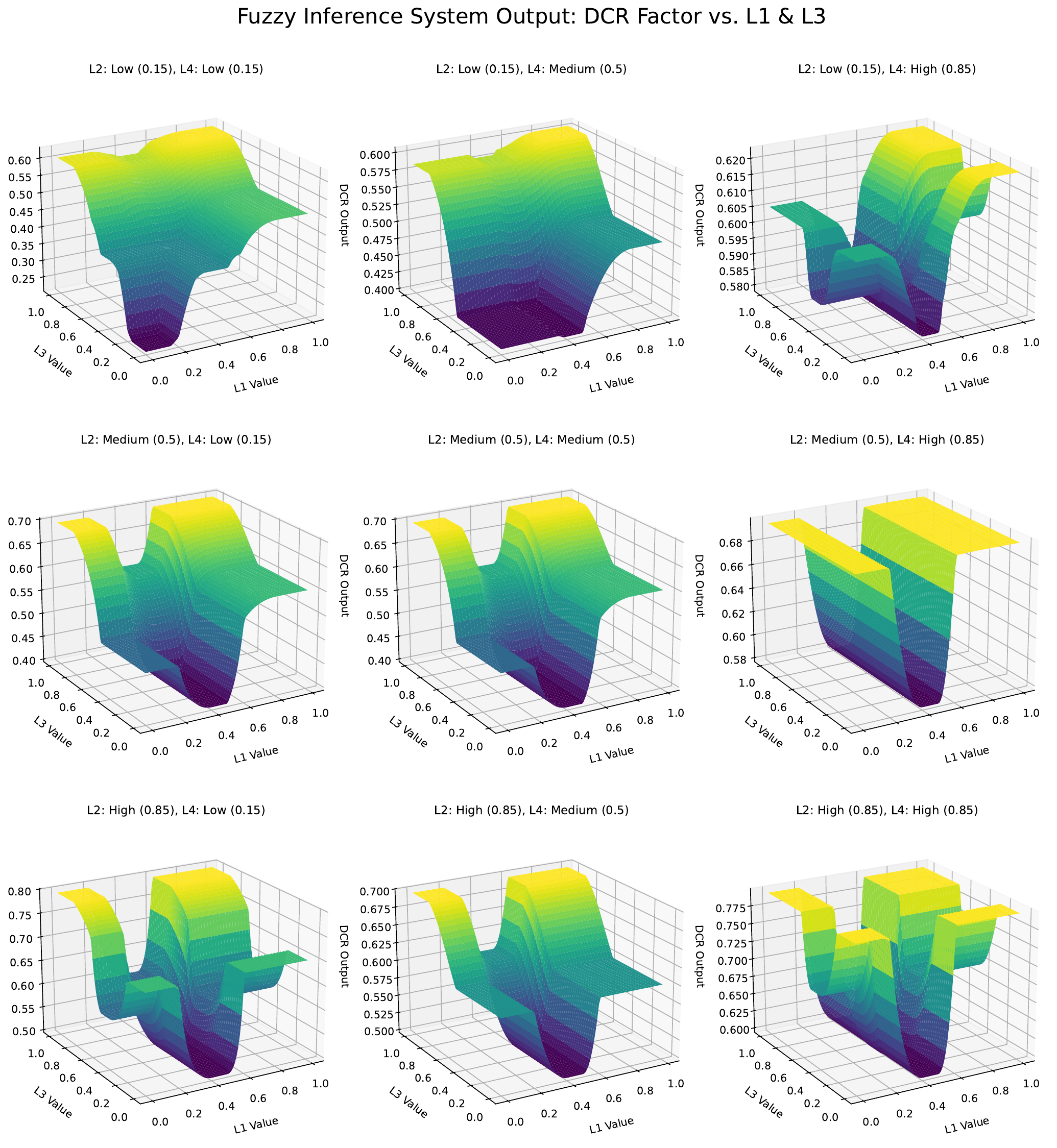}
    \caption{Fuzzy inference system output surfaces. Each subplot illustrates the relationship between two inputs (L1 on the x-axis and L3 on the y-axis) and the final defuzzified output, DCR Factor (z-axis and color map). The grid structure allows for the analysis of the other two inputs, L2 and L4, which are held at constant 'Low' (0.15), 'Medium' (0.5), and 'High' (0.85) values for each plot. Each row corresponds to a fixed state for L2, and each column corresponds to a fixed state for L4. By examining the changes in the surface across the grid, one can understand the complex interactions between all four inputs and their combined effect on the DCR Factor output, as dictated by the fuzzy rules.}
    \label{fig:fuzzyrule}
\end{figure*}

The fuzzy inference system used in the experiments is mainly implemented by the \verb|skfuzzy|\footnote{\url{https://github.com/scikit-fuzzy/scikit-fuzzy}} library. It is worth noting that we used the most intuitive settings for the fuzzy inference system in our experiments and did not carefully adjust the hyperparameters for the characteristics of the different benchmarks to achieve the best results. The purpose of doing so is to create a real evaluation setting, because in a real-world evaluation scenario, as we mentioned in section \ref{sec:motivation}, the testers will not undergo large-scale re-training just to understand the BDC risk, and thus will not have access to the "gold labels" of our experiments, and therefore, in order to avoid the possible disconnection between the DCR evaluation results and the real scenario caused by such a priori knowledge, we set up the fuzzy inference system according to simple and intuitively fixed intervals, and specifically, the design of our membership functions for fuzzy inputs and outputs are provided in Tables \ref{tab:fuzzy_membership}, their corresponding membership function visualizations are provided in Figures \ref{fig:input} and \ref{fig:output}, and the fuzzy rules for inference are provided in Table \ref{tab:fuzzy_rules}. To better understand the established fuzzy rules, we visualized them by fixing the contamination scores of L2 ($S_2$) and L4 ($S_4$) while linearly adjusting those of L1 ($S_1$) and L3 ($S_3$), the output surfaces are presented in Figure \ref{fig:fuzzyrule}. Also to take into account the perturbations caused by the randomness of the LLMs, we set a threshold value, i.e., when the DCR Factor is less than 0.02, it is considered to be generated by randomness, and it is set to 0.

\begin{table}[h]
    \centering
    \resizebox{0.48\textwidth}{!}{%
    \begin{tabular}{l|ll}
        \toprule
        \textbf{Term}  & \textbf{Function Type} & \textbf{Parameters} \\
        \midrule
        \multicolumn{3}{c}{\textbf{Fuzzy Input}} \\
        \cmidrule(l){1-3}
        Low     & Trapezoidal ($\mathtt{trapmf}$) & [0, 0, 0.1, 0.3] \\
        Medium  & Trapezoidal ($\mathtt{trapmf}$) & [0.2, 0.4, 0.5, 0.6] \\
        High    & Trapezoidal ($\mathtt{trapmf}$) & [0.5, 0.8, 1.0, 1.0] \\
        \midrule
        \multicolumn{3}{c}{\textbf{Fuzzy Output}} \\ 
        \cmidrule(l){1-3}
        Negligible  & Trapezoidal ($\mathtt{trapmf}$) & [0, 0, 0.1, 0.3] \\
        Minor       & Triangular ($\mathtt{trimf}$)    & [0.1, 0.3, 0.5] \\
        Moderate    & Triangular ($\mathtt{trimf}$)    & [0.3, 0.5, 0.7] \\
        Significant & Triangular ($\mathtt{trimf}$)    & [0.5, 0.7, 0.9] \\
        Severe      & Trapezoidal ($\mathtt{trapmf}$)  & [0.7, 0.9, 1.0, 1.0] \\
        \bottomrule
    \end{tabular}
    }
    \caption{Input \& Output Membership Functions}
    \label{tab:fuzzy_membership}
\end{table}

\begin{table}[b]
    \centering
    \resizebox{0.48\textwidth}{!}{%
    \begin{tabular}{ll}
    \toprule
    \textbf{Condition (Antecedent)} & \textbf{Output} \\
    \midrule
    L1 is Low \textbf{AND} L2 is Low \textbf{AND} L3 is Low \textbf{AND} L4 is Low  & Negligible \\[5pt]
    L3 is High \textbf{OR} L4 is High & Severe \\[5pt]
    L1 is High \textbf{OR} L2 is High & Significant \\[5pt]
    Overall medium membership is high (average of L1-L4) & Moderate \\[5pt]
    L1 is Medium \textbf{OR} L2 is Low & Minor \\
    \bottomrule
    \end{tabular}
    }
\caption{Fuzzy Rules for DCR Calculation}
\label{tab:fuzzy_rules}
\end{table}

\end{document}